# APPLYING EVOLUTIONARY OPTIMISATION TO ROBOT OBSTACLE AVOIDANCE

Olivier Pauplin, Jean Louchet, Evelyne Lutton, Michel Parent
INRIA, IMARA & COMPLEX Projects, BP 105, 78153 Le Chesnay Cedex France
olivier.pauplin@inria.fr, jean.louchet@inria.fr, evelyne.lutton@inria.fr, michel.parent@inria.fr

**Abstract:** This paper presents an artificial evolution-based method for stereo image analysis and its application to real-time obstacle detection and avoidance for a mobile robot. It uses the Parisian approach, which consists here in splitting the representation of the robot's environment into a large number of simple primitives, the "flies", which are evolved following a biologically inspired scheme and give a fast, low-cost solution to the obstacle detection problem in mobile robotics.

## 1 Introduction.

Artificial Vision, an important element in the design of autonomous robots, can be approached as the resolution of the reverse problem of reconstructing a probable model of the scene from the images. Although probabilistic optimisation methods like Evolutionary Algorithms [1],[2],[3] are in theory well adapted to the resolution of such inverse problems, their use in real applications has been relatively neglected because of their reputation of low speed and complexity. Indeed, evolving a population in which each single individual would be a complete 3-D representation of the environment should raise problems of code size and memory handling wildly out of the reach of current optimisation algorithms.

However, the technique of Parisian Evolution, introduced by Lutton et al. [4] to resolve an optimisation problem in Iterated Function Systems, showed that in some cases, splitting the representation of the object to be optimised into a collection of smaller primitives and evolve them, then use them as a collective representation of the problem's optimal solution, may lead to fast and efficient optimisation algorithms. The Fly Algorithm [5],[6] has been developed along this line to solve Computer Vision problems, using a small grain decomposition of the scene representation and evolving its components following principles inspired from Darwin's descriptions of biological evolution.

## 2 Evolutionary algorithms.

Darwin's theory assumes that a population of individuals, characterised by their genes, will evolve towards a better adaptation to its environment according to laws of natural selection. Genes mutations may occur and maintain diversity in the population.

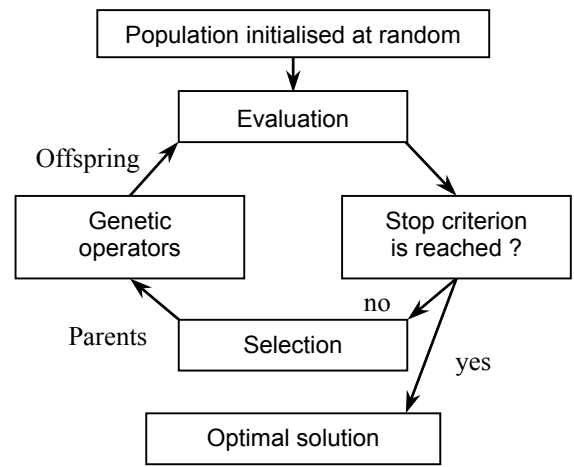

**Figure 1:** General layout of genetic algorithms.

Evolutionary algorithms manipulate individuals evaluated by a function, called fitness function, in a way similar to biological Evolution. The general diagram of such algorithms is presented in figure 1, where:
- the population is a group of individuals
- an individual is defined by his genes X = $(x_1, x_2, …, x_n)$, usually coordinates in the search space
- evaluation is the calculation of each individual's fitness value
- selection eliminates part of the population, keeping preferably the best individuals
- evolution applies genetic operators (crossover, mutations…), leading to new individuals in the population.

## 3 The Fly algorithm.

The Fly algorithm is a special case of Parisian evolution for which individuals (the "flies") are defined as 3-D points with coordinates (x, y, z). As far as we know, it is the only existing evolutionary algorithm used to detect obstacles by stereovision. The aim of the algorithm is to drive the whole population - or a significant part of it - into suitable areas of the search space, corresponding to the surfaces of visible objects in the scene.

The population of flies is initialised at random inside the intersection of two cameras' field of view. Flies then evolve following the steps of evolutionary algorithms. All cameras' calibration parameters are known.

### 3.1 Evaluation.

The fitness function used to evaluate a fly compares its projections on the left and right images given by the cameras. If the fly is on an object's surface, the projections will have similar neighbourhoods on both images and hence this fly will be attributed a high fitness.

Figures 2 and 3 illustrate that principle. Figure 3 shows neighbourhoods of two flies on left and right images. On that example, Fly1, being on an object's surface, will be given a better fitness than Fly2.

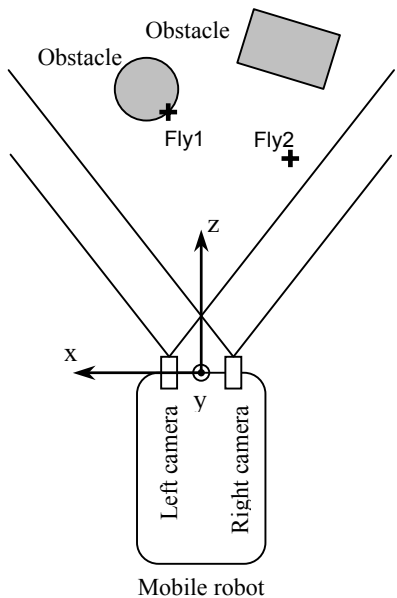

**Figure 2:** Example of device using the Fly algorithm, showing two flies from the population (top view).

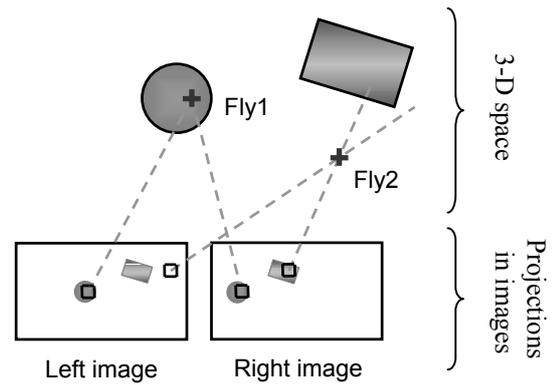

**Figure 3:** Projections of two flies in left and right images.

The mathematical expression of the fitness function is [7],[8]:

$$F = \frac{|\nabla(M_L)|.|\nabla(M_R)|}{\sum_{colours}\sum_{(i,j)\in N}[L(x_L+i, y_L+j) - R(x_R+i, y_R+j)]^2}$$

where:
- $(x_L, y_L)$ and $(x_R, y_R)$ are the coordinates of the left and right projections of the current individual
- $L(x_L+i, y_L+j)$ is the grey value at the left image at pixel $(x_L+i, y_L+j)$, similarly with R for the right image
- N is a neighbourhood around the projection of each fly, introduced to obtain a more discriminating comparison of the flies
- $|\nabla(M_L)|$ and $|\nabla(M_R)|$ are Sobel gradient norms on left and right projections of the fly. That is intended to penalise flies which project onto uniform regions, i.e. less significant flies.

### 3.2 Selection.

Selection is elitist and deterministic. It ranks flies according to their fitness values and retains the best individuals (around 40%).

A sharing operator [7],[8] reduces the fitness of flies packed together and forces them to explore other areas of the search space.

### 3.3 Genetic operators.

The following operators are applied to selected individuals.

- Barycentric cross-over: given two parents $F_1$ and $F_2$, the algorithm builds their offspring F such as:

$$\vec{OF} = \lambda \vec{OF_1} + (1-\lambda)\vec{OF_2}$$

with λ chosen at random in the interval [0,1].

- Gaussian mutation adds a Gaussian noise to each one of the three coordinates of the mutated fly. The mutation rate is set to 40%, parisian algorithms normally using a higher mutation rate than conventional evolutionary algorithms.

- Another operator, "immigration", is used to improve exploration of the search space, creating new individuals at random. It ensures a constant exploration of the search space, whose high-fitness regions evolve as the scene in front of the cameras changes.

## 4 Robot simulator.

The original way the scene is described by the population of flies led our team to adapt classical robot navigation methods in order to use the results of the Fly algorithm as input data. Boumaza [7],[9] developed a simulator of a robot moving in a simplified environment, to test theoretically control methods using the output of the Fly algorithm.

The simulator showed the possibility to build guidance methods based on the output of the Fly algorithm. Our current work consists in transferring and extending these control methods to real life situations.

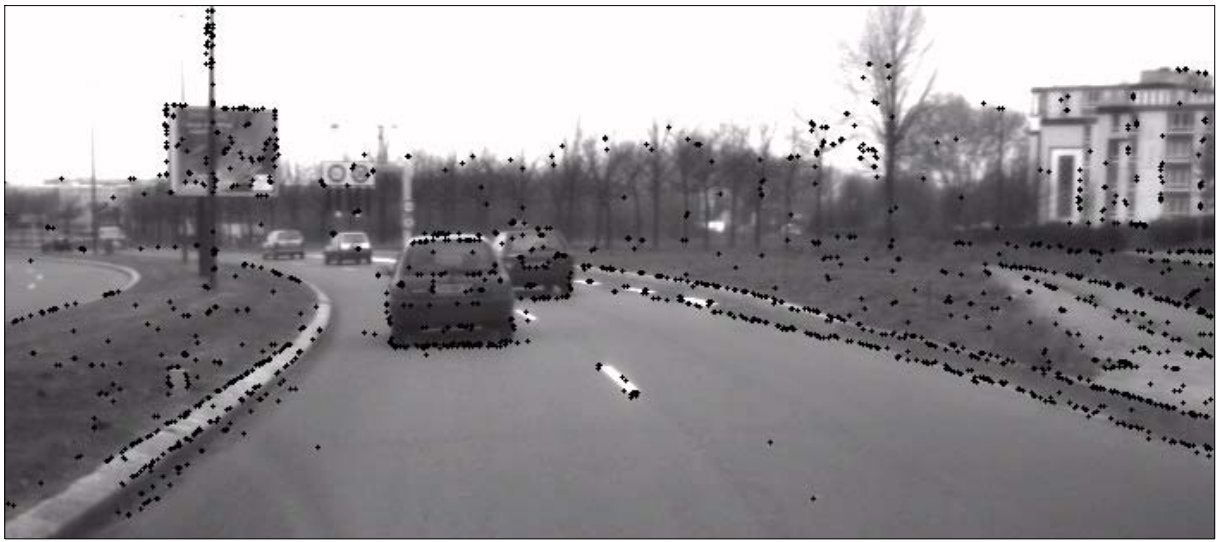

**Figure 4:** Application example of the Fly algorithm.

## 5 Real life experiments.

Figure 4 shows an application example of the Fly algorithm. Flies (black dots) concentrate on obstacles and on regions where the grey level gradient is high, for example on the roadsides. The numerator of the fitness function prevents flies from getting trapped into uniform regions (sky, road surface, etc.).

The three coordinates of each fly being known, the population of flies gives a rough description of the real 3-D scene.

### 5.1 Control.

In the scope of using the Fly algorithm in the field of automatic driving - or at least assisted driving, we developed a strategy to make the program quantify the probability that an obstacle is in front of the vehicle. The aim is to deliver a slow down or stop order when an obstacle appears close enough in the field of vision, in order to avoid frontal collision.

The general idea to achieve this goal is to see each fly as the source of a "warning value", higher when:
- the fly is near the vehicle
- the fly is in front of the vehicle (i.e. close to the z axis)
- the fly has a good fitness.

Beforehand, flies useless for this specific application have there fitness value penalised, and thus have high probability to be eliminated by the algorithm's mechanisms. We considered such flies are:
- flies more than 2 metres above the road surface
- flies with a height under 10 centimetres (detecting the ground)
- flies more than 16 metres ahead of the vehicle.

An experimental analysis led us to choose the simple following formula for the warning value of a fly:

$$warning(fly) = \frac{F}{x^2 \times z}$$

where F is the fitness value of the fly, and z and x its coordinates as shown on figure 2.

For |x| < 0.5 m we consider x = 0.5 m, and for z < 1 m we consider z = 1 m. This is to avoid giving excessive warning values to flies with a not necessarily good fitness but with a very small x or z coordinate. Moreover, obstacles within a range of half a metre to the left or to the right from the centre of the vehicle (|x| < 0.5 m) are equally dangerous, and are consequently processed the same way.

The warning function was built in order to give high warning values to flies for which the three coefficients F, $1/x^2$ and $1/z$ are simultaneously high. Indeed a fly with a low fitness value (thus probably not on an obstacle), far from the vehicle or not in front of it, does not show evidence of an imminent collision. Experiments with a $1/x$ factor instead of $1/x^2$ did not give satisfactory results, as it tended to overestimate the importance of flies off the cameras axis.

### 5.2 Results.

To validate the algorithm, we tested it on two stereo pairs of images: one representing a road with no immediate obstacle (figures 5 and 6), and one representing a pedestrian crossing the street in front of the vehicle (figures 7 and 8). Figure 5 does not show a case of emergency breaking, whereas figure 7 shows a situation closer to a collision.

Results are obtained using two commercial CCD cameras and a computer (Pentium 2 GHz). The population of flies is 5000. One generation takes about 10 milliseconds. Population update and calculation of the warning values are done in a quasi-continuous way, and the system needs about 10 to 30 generations to react to a new event in the scene.

Figures 5 and 7 show the 250 best flies of the resulting population. Flies appear as black crosses. We note that, on both figures 5 and 7, flies gather on the visible objects of the scene (car, pedestrian, road sides...).

Figures 6 and 8 show the same (x,y) view as figures 5 and 7, with only flies represented. Flies appear as spots as dark as their warning value is high.

We note the algorithm delivers higher warning values in figure 8 than in figure 6, where they are very close to zero.

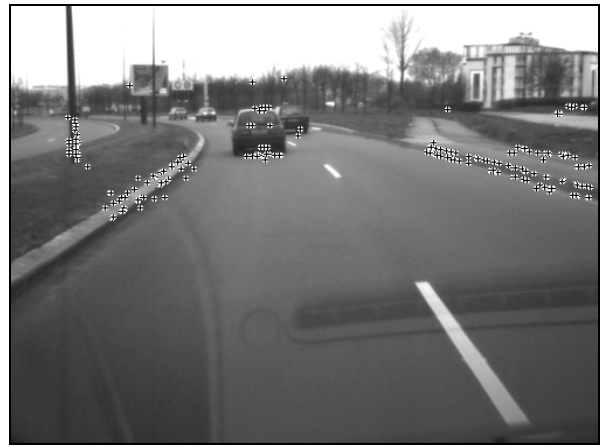

**Figure 5:** A road with no immediate obstacle.

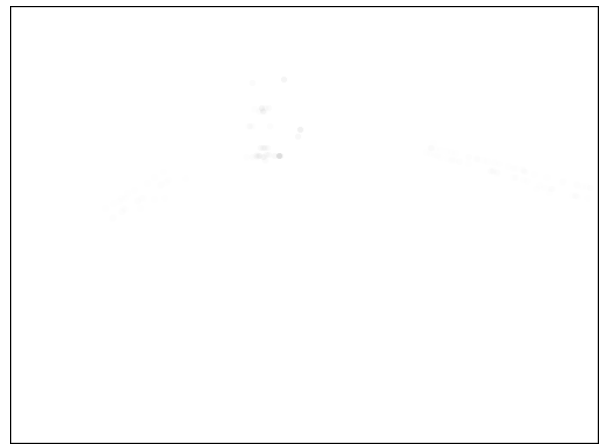

**Figure 6:** Warning values of figure 5 flies.

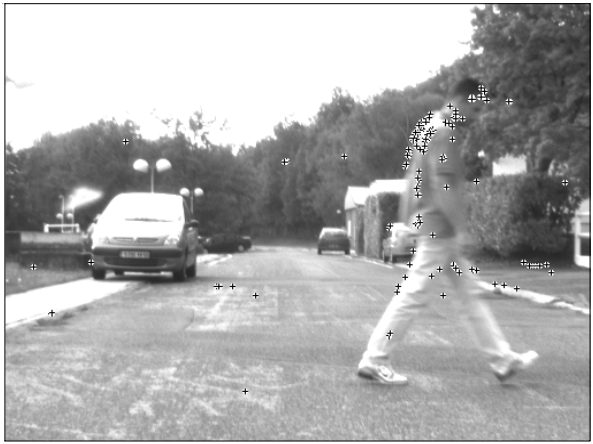

**Figure 7:** A pedestrian at 4 metres from the cameras, on the middle of the road.

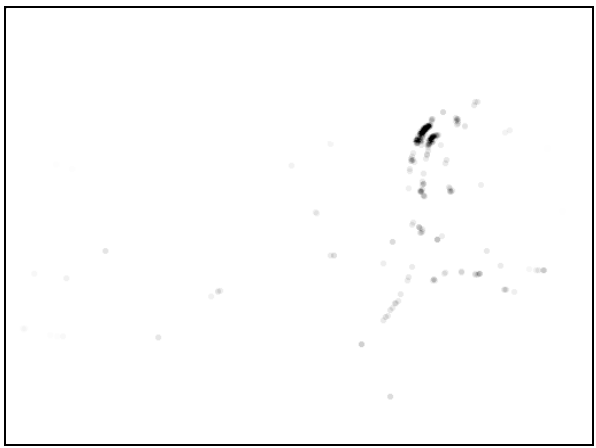

**Figure 8:** Warning values of figure 7 flies.

A global warning value can be defined as the mean of the warning values of a population. In the first case, this mean is 0.09, whereas in the second case it is 0.85. The high difference between these two values suggests that they can be used to discriminate between the two situations. Further experiments will be needed in order to confirm or refine this criterion.

## 6  Conclusion.

The Fly algorithm has proved a valid method for obstacle detection in outdoor environments. The simplicity of the fitness function used opens the way to real time applications. Real time vehicle control based on the information of flies (coordinates, fitness value) has been developed.

Classical image segmentation and stereo reconstruction methods are potentially able to give more complete and accurate results than the Fly algorithm, though requiring higher processing times. However, the Fly algorithm presents some features which are outstandingly interesting in real time vision applications: in particular its asynchronous properties and its principle of continuous refinement of previous results, giving reaction times to new events intrinsically faster than classical methods [8].

Our future work will be directed toward developing guidance algorithms for mobile robots in real life situations, and to integrate them into a vehicle of IMARA project.

## Acknowledgements.

We thank Dr Amine Boumaza for his important contribution to the development of the code used in our experiments.

This research was funded in part by the IST Programme of the European Commission in the CyberCars project:
http://www.cybercars.org/

## References.